\documentclass[runningheads]{llncs}
\usepackage{times}
\usepackage{epsfig}
\usepackage{graphicx}
\usepackage{amsmath}
\usepackage{amssymb}
\usepackage{cite}
\usepackage{booktabs}
\usepackage{hyperref}
\usepackage{multirow}
\usepackage{caption}
\usepackage{subcaption}
\usepackage{tablefootnote}
\usepackage{multirow}
\usepackage{tablefootnote}
\usepackage{threeparttable}
\usepackage{float}
\usepackage{xcolor} 
\usepackage{array}
\graphicspath{ {./images/} }
\newcolumntype{P}[1]{>{\centering\arraybackslash}p{#1}}
\AtBeginDocument{\hypersetup{pdfborder={0 0 1}}}
\begin{document}

\title{IndicSTR12: A Dataset for Indic Scene Text Recognition}

\author{Harsh Lunia\orcidID{0009-0007-4155-2011} \and
Ajoy Mondal\orcidID{0000-0002-4808-8860} \and
C. V. Jawahar\orcidID{0000-0001-6767-7057}}
%
\institute{Centre for Vision Information Technology \\
International Institute of Information Technology, Hyderabad - 500032, INDIA
\url{http://cvit.iiit.ac.in/research/projects/cvit-projects/indicstr}
\email{harsh.lunia@research.iiit.ac.in}
\email{\{jawahar,ajoy.mondal\}@iiit.ac.in}}

\maketitle              

\begin{abstract}

The importance of Scene Text Recognition (STR) in today's increasingly digital world cannot be overstated. Given the significance of STR, data-intensive deep learning approaches that auto-learn feature mappings have primarily driven the development of STR solutions. Several benchmark datasets and substantial work on deep learning models are available for Latin languages to meet this need. On more complex, syntactically and semantically, Indian languages spoken and read by 1.3 billion people, there is less work and datasets available. This paper aims to address the Indian space's lack of a comprehensive dataset by proposing the largest and most comprehensive real dataset - IndicSTR12 - and benchmarking STR performance on 12 major Indian languages~\footnote{Assamese, Bengali, Odia, Marathi, Hindi, Kannada, Urdu, Telugu, Malayalam, Tamil, Gujarati, and Punjabi}. A few works have addressed the same issue, but to the best of our knowledge, they focused on a small number of Indian languages. The size and complexity of the proposed dataset are comparable to those of existing Latin contemporaries, while its multilingualism will catalyse the development of robust text detection and recognition models. It was created specifically for a group of related languages with different scripts. The dataset contains over 27000 word-images gathered from various natural scenes, with over 1000 word-images for each language. Unlike previous datasets, the images cover a broader range of realistic conditions, including blur, illumination changes, occlusion, non-iconic texts, low resolution, perspective text etc. Along with the new dataset, we provide a high-performing baseline on three models: PARSeq (Latin SOTA), CRNN, and STARNet.

\end{abstract}

\section{Introduction}

Language has enabled people world around to exchange and communicate. Different communities have recognized the importance of the same, which becomes evident by the diversity of languages across the globe. The textual representation of language, on the other hand, significantly broadens the scope of transfer. Semantically rich writing found in the wild has powerful information that can substantially aid in understanding the surrounding environment in the modern era. Textual information found in the wild is used for various tasks, including image search, translation, transliteration, assistive technologies (particularly for the visually impaired), autonomous navigation, and so on. The issue of automatically reading text from photographs or frames of a natural environment is referred to as Scene Text Recognition (STR) or  Photo-OCR. This problem is typically subdivided into two sub-problems: Scene text detection, which deals with locating text within a picture, and cropped word image recognition. Our work addresses the second sub-problem: recognizing the text in a clipped word image.

Traditionally, OCR has focused on reading printed or handwritten text in documents. However, as capturing devices such as mobile phones and video cameras have proliferated, scene text recognition has become critical and a problem whose solution holds promise for furthering the resolution of other downstream tasks. Although there has been considerable progress in STR, it has some unique issues, i) varying backgrounds in natural scenes, ii) varying script, font, layout, and style, and iii) text-related image flaws such as blurriness, occlusion, uneven illumination, etc. Researchers have attempted to address the aforementioned issue by amassing datasets specifically tailored to a given problem, each of which has some distinguishing features and represents a subset of challenges encountered in real-world situations. For example,~\cite{phan2013recognizing} has perspective text,~\cite{wang2011end} has blurred text, and~\cite{risnumawan2014robust} has curved text.

Rather than designing and testing manually created features, nearly all current solutions rely on deep learning techniques to automate feature learning. Because of the data-intensive nature of these models, it has become standard practice to train the models on synthetically generated data that closely resemble real-world circumstances and test the trained models on difficult-to-obtain real datasets. STR solutions for Latin languages such as English have made significant progress. However, Latin STR having reached a certain level of maturity, has begun to train solely on available real datasets~\cite{baek2021if} to achieve nearly comparable performance compared to a mix of synthetic and real. Using an already available diverse set of public real datasets totaling almost 0.3 million image instances for English was a significant factor that led to similar results.

Because Indian language scripts are visually more complex, and their output space is much larger than English languages, not all Latin STR models can mimic performance in Indian STR solutions~\cite{mathew2016multilingual}. STR solutions have not progressed in the case of Indian languages, which are spoken by 18\% of the world population, due to a lack of real datasets and models that are better equipped to handle the inherent complexities of the languages. Non-Latin languages have made less progress, and existing Latin STR models need to generalize better to different languages~\cite{chen2021text}.
\begin{figure}
    \centering
    \includegraphics[width=\textwidth,keepaspectratio]{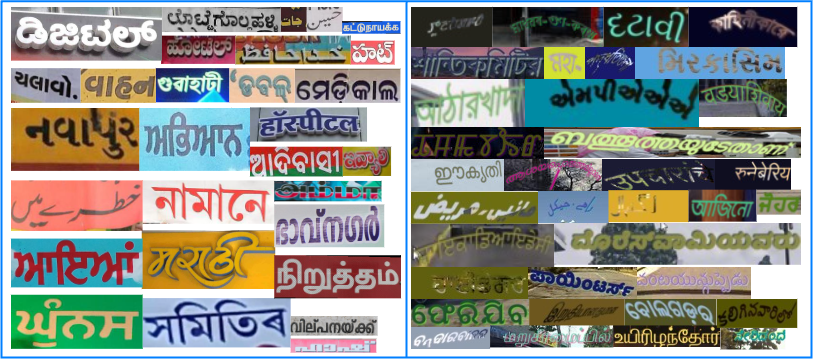}
    \caption{Samples from IndicSTR12 Dataset: Real word-images (left); Synthetic word-images (right)}
    \label{fig:language_samples}
\end{figure}

\subsubsection{Contribution}
Given the need for more data for non-Latin languages, particularly Indian languages, our work attempts to address the issue by contributing as follows:

\begin{enumerate}
\item We propose a real dataset (Fig.~\ref{fig:language_samples} (left)) for 12 Major Indian Languages, namely - Assamese, Bengali, Odia, Marathi, Hindi, Kannada, Urdu, Telugu, Malayalam, Tamil, Gujarati and Punjabi - wherein Malayalam, Telugu, Hindi, and Tamil word-images have been taken from~\cite{8270315} and~\cite{gunna2021transfer}. Since the number of word instances proposed by~\cite{gunna2021transfer} for Gujarati was less than 1000 work image instances, we augment the proposed Gujarati instances to achieve numbers comparable to other languages in the proposed dataset.

\item We propose a synthetic dataset for all 13 Indian languages (Fig.~\ref{fig:language_samples} (right)), which will help the STR community progress on multi-lingual STR, in effect similar to SynthText~\cite{gupta2016synthetic} and MJSynth~\cite{jaderberg2014synthetic}.

\item Finally, we compare the performance of three STR models - PARSeq~\cite{bautista2022parseq}, CRNN, and STARNet~\cite{liu2016star} - on all 12 Indian languages, some of which have no previous benchmark to compare with. The effectiveness of these models on the IndicSTR12 dataset and other publicly accessible datasets supports our dataset's claim that it is challenging (Table~\ref{table:otherDataset}). 

\item By simultaneously training on multiple languages' real datasets, we demonstrate how multi-lingual recognition models can aid models in learning better, even with sparse real data.
\end{enumerate}

\section{Related Works}
 
Scene text recognition (STR) models use CNNs to encode image features. For decoding text out of the learnt image features in a segmentation freeway, it relies either on Connectionist Temporal Classification (CTC)~\cite{graves2006connectionist} or encoder-decoder framework~\cite{sutskever2014sequence} combined with attention mechanism~\cite{bahdanau2014neural}. The CTC-based approaches~\cite{su2015accurate,liu2016star} treat images as a sequence of vertical frames and combine prediction per frame based on a rule to generate the whole text. In contrast, the encoder-decoder framework~\cite{liu2018squeezedtext} uses attention to align input and output sequences. There has been work on both CTC and attention-based models for STR. DTRN~\cite{he2016reading} is the first to use CRNN models, a combination of CNNs with RNNs stacked on them, to generate convolutional feature slices to be fed to RNNs. Using attention,~\cite{liu2018squeezedtext} performs STR based on encoder-decoder model wherein the encoder is trained in binary constraints to reduce computation cost. Work on datasets of varying complexity has been done to promote research on STR for challenging scenarios as well. A few challenging ones in terms of occlusion, blur, small and multi orientation word-images are ICDAR 2015~\cite{karatzas2015icdar}, Total-Text~\cite{ch2020total}, LSVT~\cite{sun2019chinese, sun2019icdar}.

\subsubsection{Indian Scene Text Recognition}

The lack of annotated data is a hurdle, especially in the case of Indian languages, in realizing the success achieved in the case of Latin STR solutions. There has been attempt to address the data scarcity problem over the years in a scattered and very language-specific way for Indian languages. \cite{chandio2020cursive} is the first work to propose an Urdu dataset and benchmark STR performance on Urdu text. It contains 2,500 images, giving 14,100 word-images. The MLT-17 dataset~\cite{nayef2017icdar2017} contains 18k scene images in multiple languages, including Bengali. Building on top of it, MLT-19~\cite{nayef2019icdar2019} contains 20k scene images in multiple languages, including Bengali and Hindi. To our knowledge, this is currently the only multilingual dataset, and it supports ten different languages.~\cite{8270315} trains a CRNN model on synthetic data for three Indian languages: Malayalam, Devanagari (Hindi), and Telugu. It also releases an IIIT-ILST dataset for mentioned three languages for testing, reporting a WRR of 42.9\%, 57.2\% and 73.4\% in Hindi, Telugu, and Malayalam, respectively. \cite{buvsta2019e2e} proposes a CNN and CTC-based method for script identification, text localization, and recognition. The model is trained and tested on MLT 17 dataset, achieving 34.20\% WRR for Bengali. An OCR-on-the-go model~\cite{saluja2019ocr} obtained a WRR of 51.01\% on the IIIT-ILST Hindi dataset and a CRR of 35\% on a multi-lingual dataset containing 1000 videos in English, Hindi, and Marathi. \cite{gunna2021transfer} explored transfer learning among Indian languages as an approach to increase WRR and proposed a dataset of natural scene images in Gujarati and Tamil to test the hypothesis further. It achieved a WRR gain of 6, 5 and 2\% on the IIIT-ILST dataset and a WRR of 69.60\% and 72.95\% in Gujarati and Tamil respectively. 

\section{Datasets and Motivation}

\subsection{Synthetic Dataset}
\label{Dataset_synth}

\begin{table}
\centering
\caption{Statistics of Synthetic Data}
\label{table:SynthStats}
\begin{tabular}{P{2.5cm}|P{2.0cm}|P{1.8cm}|P{1.5cm}|P{1.5cm}}\hline
Language          &Vocabulary Size &Mean, Std &Min, Max &Fonts  \\\hline\hline
Gujarati          &106,551 &5.96 , 1.85  &1 , 20 &12 \\
Urdu              &234,331 &6.39 , 2.20  &1 , 42 &255 \\
Punjabi           &181,254 &6.45 , 2.18  &1 , 31 &141 \\
Manipuri (Meitei) &66,222  &6.92 , 2.49  &1 , 29 &24  \\
Assamese          &77,352  &7.08 , 2.70  &1 , 46 &85 \\
Odia              &149,681 &8.00 , 3.00  &1 , 37 &30 \\
Bengali           &449,986 &8.53 , 3.00  &1 , 38 &85 \\
Marathi           &180,278 &8.64 , 3.43  &1 , 44 &218 \\
Hindi             &319,982 &8.76 , 3.20  &1 , 50 &218 \\
Telugu            &499,969 &9.75 , 3.38  &1 , 50 &62  \\
Tamil             &399,999 &10.75 , 3.64 &1 , 35 &158 \\
Kannada           &499,972 &10.72 , 3.87 &1 , 41 &30  \\
Malayalam         &320,000 &14.30 , 5.36 &1 , 53 &101 \\ \hline
\end{tabular}
\end{table}

It has been an accepted practice in the community to train an STR model on a large synthetically generated dataset since~\cite{jaderberg2014synthetic} trained a model on 8 Million synthetically generated English word-images called MJSynth. This trend, as~\cite{baek2019wrong} points, is due to the high cost of annotating real data. Another synthetic dataset widely in use for English language is SynthText~\cite{gupta2016synthetic}. On the Indian language side, there have been some works like~\cite{8270315,gunna2021transfer} which use synthetic datasets for a total of 6 Indian Languages. However, like real dataset scenario, a comprehensive synthetic dataset for all 12 major Indian languages is absent. We extend the previously referred work and propose a synthetic dataset for all 12 Major Indian languages by following the same procedure as~\cite{8270315}. The word images are rendered by randomly sampling words from a vocabulary of more than 100K words for each language (except for Assamese) and rendering them using freely available Unicode fonts~\ref{table:SynthStats}. Each word image is first rendered in the foreground layer by varying font, size, and stroke thickness, color, kerning, rotation along the horizontal line, and skew. This is followed by the applying a random perspective projective transformation to the foreground layer and, consequently, a blending of the same with a random crop from a natural scene image taken from Places365 dataset~\cite{7968387}. Lastly, the foreground image is alpha composed with a background image which can either be a random crop from a natural scene image or one having a uniform color. The synthetic dataset proposed has more than 3 Million word images per language. For benchmarking STR performance, we have followed the same procedure as~\cite{gunna2021transfer}, using 2 Million word images for training the network and 0.5 Million for validation and testing.

\subsection{Real Dataset}\label{Dataset_real}

\begin{table}[h]
\label{table:LeftOutLangs}
\centering
\caption{Usage Statistics and General Information of Official Indian Languages not part of IndicSTR12 Dataset}
\begin{tabular}{P{2.0 cm}|P{3.5 cm}|P{1.5 cm}|P{2.0 cm}}
\hline
Language  &Script       &Usage  &Family     \\\hline\hline
Bodo                            &Devanagari                  &1.4M &Sino-Tibetan \\ 
Kashmiri      & Arabic \& Devanagari &11M   &Indo-Aryan \\ 
Dogri         &Devanagari  &2.6M &Indo-Aryan \\ 
Konkani       &Devanagari  &2.3M &Indo-Aryan \\ 
Maithali      &Devanagari  &34M  &Indo-Aryan \\ 
Sindhi        & Arabic \& Devanagari &32M   &Indo-Aryan \\ 
Santhali      &Ol Chiki    &7.6M  &Austroasiatic \\ 
Nepali        &Devanagari  &25M   &Indo-Aryan    \\\hline
\end{tabular}
\end{table}

According to the Census 2011 report on Indian languages~\cite{IndiaCensus2011}, India has 22 major or scheduled languages with a significant volume of writing. All major Indian languages can be classified into four language families: Indo-Aryan, Dravidian, Sino-Tibetan, and Austro-Asian (listed in decreasing order of usage). Sanskrit, Bodo, Dogri, Kashmiri, Konkani, Maithili, Nepali, Santali, and Sindhi are among the languages not covered by the IndicSTR12 dataset~\hyperref[table:LeftOutLangs]{Table 2}. As a classical language, Sanskrit has a long history of heavily influencing all of the subcontinent's languages. It is now widely taught at the secondary level, but its use is limited to ceremonial and ritualistic purposes with no first-language speakers. Other languages that have been left out either have scripts that are similar to one of the included languages~\footnote{Bodo, Dogri, Kashmiri, Konkani, Maithili, Nepali, and Sindhi} or have minimal usage in the domain of scene text in natural settings~\footnote{Santali}. According to the 2011 census report~\cite{IndiaCensus2011}, the included languages cover 98\% of the subcontinent's spoken language. IndicSTR12 is an extension of IIIT-ILST~\cite{8270315} and~\cite{gunna2021transfer}, which cover Telugu, Malayalam, Hindi, Gujarati, and Tamil, respectively. There has been no addition of images for any of the mentioned languages, except for Gujarati, which had less than 1000 word-images.

\subsubsection{IndicSTR12 Curation Details}
\begin{figure}[h]
\centering
\includegraphics[width=\textwidth,keepaspectratio]{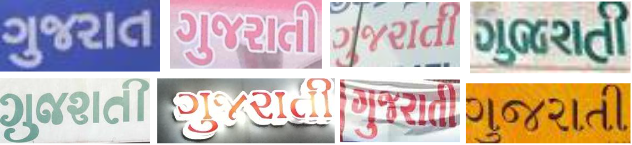}
\caption{\textbf{IndicSTR12 Dataset}: Font Variations for the same word - Gujarati or Gujarat}
\label{fig:font_variation}
\end{figure}

\begin{figure}[h]
\centering
\includegraphics[width=\textwidth,keepaspectratio]{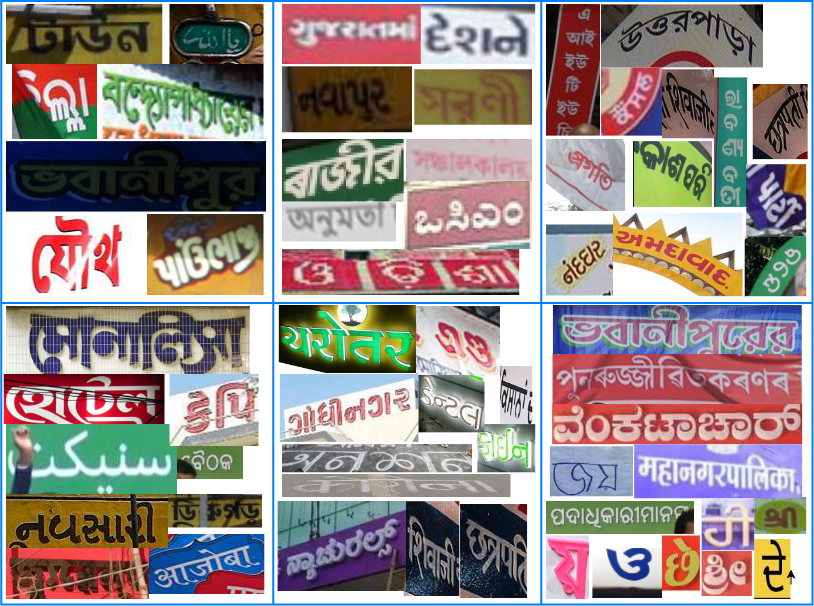}
\caption{IndicSTR12 Dataset Variations, clockwise from Top-Left: Illumination variation, Low Resolution, Multi-Oriented - Irregular Text, Variation in Text Length, Perspective Text, and Occluded.}
\label{fig:diversity_quality}
\end{figure}

All the images have been crawled from Google Images using various keyword-based searches to cover all the daily avenues wherein Indic language text can be observed in natural settings. To mention some - Wall paintings, railway stations, Signboards, shop/temple/mosque/gurudwara name-boards, advertisement banners, political protests, house plates, etc. Because they have been crawled from a search engine, they come from various sources, offering a wide range of conditions under which images were captured. Curated images have blur, non-iconic/iconic text, low-resolution, occlusion, curved text, perspective projections due to non-frontal viewpoints, etc~\ref{fig:diversity_quality} and~\ref{fig:font_variation}.

\subsubsection{IndicSTR12 Annotation Details}
All the words within the image are annotated with four-corner point annotation as done in~\cite{nayef2019icdar2019} to capture both horizontal and curved word structures of scene texts. The annotators were encouraged to follow the reading direction and include as little background space as possible. To further ensure the quality of annotation, all the annotated data was reviewed by another entity to ensure proper removal/correction of empty or wrong labels. The reviewers were also tasked with further classifying each word instance into the three categories of Oriented Text, Low-resolution/Smaller Text, and Occluded Text. This will enable community members to assess which areas of a model's performance require special consideration. There are at least 1000 word images per language and their corresponding labels in Unicode. The dataset can also be used for the problem of script identification and scene text detection. 

\subsection{Comparision With Existing Datasets}\label{Comparision_Public_STR}
\begin{table}
\centering
\caption{Statistics of Various Public STR Real Dataset}
\label{table:Stats_Public_STR}
{\renewcommand{\arraystretch}{1.3}%
\begin{tabular}{l|c|l|l|c}\hline 
Dataset   &Word Images &Language &Features &Tasks\tablefootnote{'D' Stands for Detection and 'R' for Recognition}  \\
& (train/test) & & & \\ \hline\hline 
IIIT5K-Words~\cite{mishra2012scene}    &2K/3K &English &Regular  &R \\
SVT~\cite{wang2011end}  &211/514  &English &Regular, blur, low resolution  &D,R \\
ICDAR2003~\cite{lucas2005icdar} &1157/1111 &English &Regular &D,R \\
ICDAR2013~\cite{karatzas2013icdar} &3564/1439 &English &Regular, Stroke Labels &D,R \\
ICDAR2015~\cite{karatzas2015icdar} &4468/2077 &English &Irregular, Blur, Small &D,R \\ 
SVT Perspective~\cite{phan2013recognizing} &0/639 &English &Irregular, Perspective Text &R \\
MLT-19~\cite{nayef2019icdar2019} &89K/102K & Multi-Lingual~\tablefootnote{Arabic, Bangla, Chinese, Devanagari, English, French, German, Italian, Japanese, and Korean} &Irregular &D,R \\ 
MTWI~\cite{he2018icpr2018} &141K/148K &Chinese, English &Irregular &R \\
LSVT~\cite{sun2019chinese,sun2019icdar} &30K/20K& Chinese, English &Irregular, multi-oriented & D, R \\
MLT-17~\cite{nayef2017icdar2017} &85K/11K &Multi-Lingual\tablefootnote{Arabic, Bangla, Chinese, English, French, German, Italian, Japanese, and Korean}  &Irregular &D,R \\
Urdu-Text~\cite{chandio2020cursive} &14100 &Urdu &Irregular, Noisy &D,R \\
CUTE80~\cite{risnumawan2014robust} &0/288 &English &Irregular, Perspective Text, &D,R \\
&    &  &Low Resolution & \\\hline 
IndicSTR12 (Ours)~\tablefootnote{Extends IIIT-ILST~\cite{8270315,gunna2021transfer}}  &20K/7K &Multi-Lingual~\tablefootnote{Assamese, Bengali, Odia, Marathi, Hindi, Kannada, Urdu, Telugu, Malayalam, Tamil, Gujarati and Punjabi} &Irregular, Low Resolution, &D,R \\
&  & &Blur, Occlusion, Perspective Text  & \\\hline
\end{tabular}}
\end{table}

Real datasets were initially utilised for fine-tuning models trained on synthetic datasets and evaluating trained STR models because the majority of real data sets only contain thousands of word-images. There are many datasets available for English Languages that address a variety of difficulties, but the community has only now begun to consider training models with real data~\cite{baek2021if}. Broadly a real dataset can be seen as regular and irregular types. Regular datasets have most word-images that are iconic (frontal), horizontal and a small portion of distorted samples. In contrast, the majority of the word-images in irregular datasets are perspective text, low-resolution, and multi-oriented. The high variation in text instances makes these difficult for STR. For examples and more information, please refer to Table~\ref{table:Stats_Public_STR}. Our Dataset has been curated with the goal of catering to both regular and irregular samples. Being extracted from the Google search engine via various keywords generally associated with scene and text, it tries to cover a wide range of natural scenarios, mostly seen in Indian Scene texts. The classification of word-images into categories was done to allow Indian STR solutions to assess their current standing on regular texts and, as the solutions advance, to provide a challenging subset to further refine the prediction models.

\section{Models}\label{models_benchmark}

This section explains the models used to benchmark STR performance on 12 Indian languages in IndicSTR12 dataset. Three models were picked up to benchmark the STR performance - PARSeq~\cite{bautista2022parseq} is the current state-of-the-art Latin STR model, CRNN~\cite{shi2016end} which has low accuracy than a lot of current models but is widely chosen by the community for practical usage because it's lightweight and fast. Another model called STARNet~\cite{liu2016star}, extracts more robust features from word-images and performs an initial distortion correction, is also used for benchmarking. This model has been taken up to maintain consistency with the previous works on Indic STR~\cite{8270315,gunna2021transfer}.

\subsubsection{PARSeq}

\begin{figure}[h]
\centering
\includegraphics[width=\textwidth,keepaspectratio]{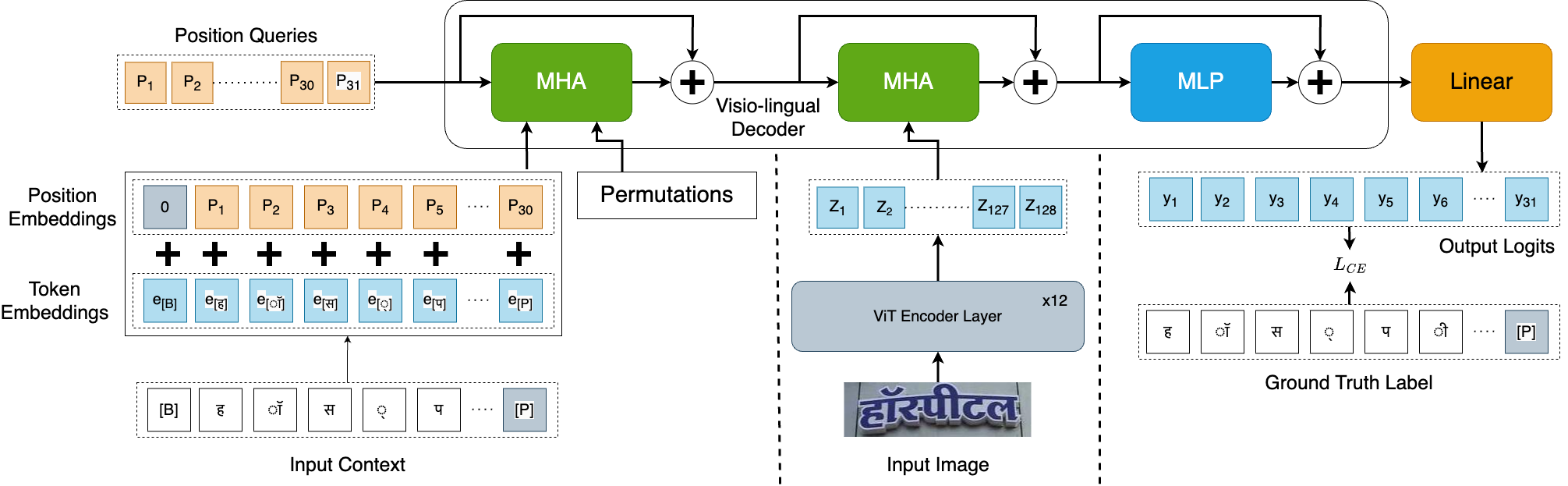}
\caption{PARSeq architecture. [B] and [P] begin the sequence and padding tokens. T=30 or 30 distinct position tokens. \(L_{CE}\) corresponds to cross entropy loss.}
\label{fig:parseq}
\end{figure}

PARSeq is a transformer-based model which is trained using Permutation Language Modeling (PLM). Multi-head Attention~\cite{NIPS2017_3f5ee243} is extensively used, MHA(q, k, v, m), where q, k, v and m refer to query, key, value and optional attention mask. 
The model follows an encoder-decoder architecture wherein the encoder stack has 12 encoder blocks while the decoder has only a single block. 

The model uses 12 vision transformer encoder blocks each containing 1 self-attention MHA module. The image \(x \in \mathbb{R^{W\times H\times C}}\) is tokenized evenly into \(p_w\times p_h\) patches which are in-turn projected into \(d_{model}\) - dimensional tokens using an embedding matrix \(\boldsymbol{W}^{p} \in \mathbb{R}^{p_{w} p_{h} C \times d_{model}}\). Position embeddings are then added to tokens before sending them to the first ViT encoder block. All output tokens \(\boldsymbol{z}\) are used as input to the decoder: 

\[\boldsymbol{z} = Enc(x) \in \mathbb{R}^{\frac{WH}{p_wp_h}\times d_{model}}\]

The Visio-lingual Decoder used is a pre-layerNorm transformer decoder with two MHAs. The first MHA requires position tokens, \(\boldsymbol{p} \in \mathbb{R}^{(T + 1) \times d_{model}}\) (\(T\) being context length), context embeddings,  \(\boldsymbol{c} \in \mathbb{R}^{(T + 1) \times d_{model}}\), and attention mask,  \(\boldsymbol{m} \in \mathbb{R}^{(T + 1) \times (T + 1}\). The position token captures the target position to be predicted and decouples the context from the target position, enabling the model to learn from permutation language modeling. The attention masks vary at different use points. During training, they are based on permutations, while during inference it is a left-to-right look-ahead mask. Transformers process all tokens in parallel, therefore to enforce the condition of past tokens having no access to future ones, attention masks are used. PLM, which in theory requires the model to train on all \(T!\) factorizations, in practice is achieved by using attention masks to enforce some subset \(K\) of \(T!\) permutations.

\[\boldsymbol{h}_{c} = \boldsymbol{p} + MHA(\boldsymbol{p}, \boldsymbol{c}, \boldsymbol{c}, \boldsymbol{m}) \in \mathbb{R}^{(T + 1) \times d_{model}}\]

The second MHA is used for image-position attention where no attention mask is used.
\[\boldsymbol{h}_i = \boldsymbol{h}_c + MHA(\boldsymbol{h}_c, \boldsymbol{z}, \boldsymbol{z}) \in \mathbb{R}^{(T + 1) \times d_{model}}\]

The last decoder hidden state is used to get the output logits \(\boldsymbol{y} = Linear(\boldsymbol{h}_{dec} \in \mathbb{R}^{(T + 1) \times (S + 1)}\) where S is the size of the character set and an addition of 1 is due to end of sequence token \(\boldsymbol{[E]}\).

The decoder block can be represented by: 

\[\boldsymbol{y} = Dec(\boldsymbol{z}, \boldsymbol{p}, \boldsymbol{c}, \boldsymbol{m}) \in \mathbb{R}^{(T + 1) \times (S + 1)}\]

\subsubsection{CRNN}
\begin{figure}[h]
\centering
\includegraphics[width=\textwidth,keepaspectratio]{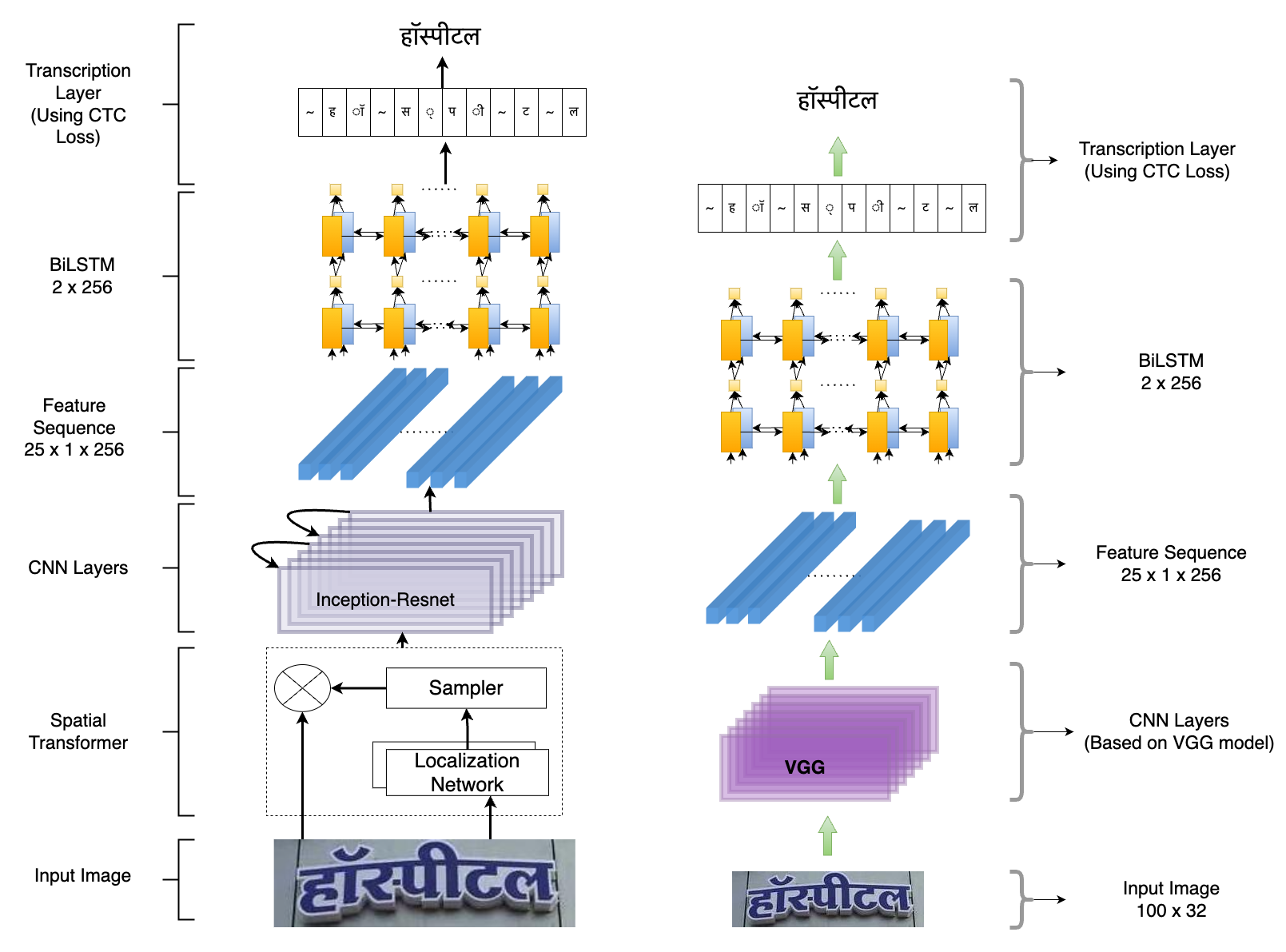}
\caption{STARNet model (left) and CRNN model (right)}
\label{fig:CRNN}
\end{figure}

CRNN is a combination of CNN and RNN, as shown in Fig.~\ref{fig:CRNN} (right). The model primarily can be viewed as a combination of 3 components - (i) an encoder, here a standard VGG model~\cite{simonyan2014very}, to extract features from word-image, (ii) a decoder consisting of RNN and lastly, (iii) a Connectionist Temporal Classification (CTC) layer which aligns decoded sequence to ground truth. The CNN encoder is made of 7 layers to extract the feature maps. For RNN, it uses a two-layer BiLSTM model, each with a hidden size of 256 units. During training, the CTC layer provides non-parameterized supervision to ensure that predictions match the ground truth. All of our experiments used a PyTorch implementation of the model by~\cite{shi2016end}.

\subsubsection{STARNet}

STARNet in Fig.~\ref{fig:CRNN} (left), like CRNN, consists of three components: an encoder that is a CNN-based model, a RNN-based decoder, and a Connectionist Temporal Classification (CTC) layer to align the decoded sequence with the ground truths. However, it differs from CRNN in two ways: it performs initial distortion correction using a spatial transformer, and its CNN is based on an inception ResNet architecture~\cite{szegedy2017inception}, which can extract more robust features required for STR.

\section{Experiments}\label{experiments_models}

Each STR model is trained on 2 million and tested on 0.5 million synthetic word-images. To further adapt the model to real-world word-images, it is trained and tested on the proposed IndicSTR12. We use 75\% of the word-images for training and 25\% for testing in each language. 

All PARSeq models are trained on dual-GPU platforms with Pytorch DDP for 20-33 epochs and 128 batch size. In conjunction with the 1cycle learning rate scheduler~\cite{smith2019super}, the Adam optimizer~\cite{kingma2014adam} is used. As in the PARSeq model, we use K = 6 permutations with mirroring for PLM and an 8 x 4 patch size for ViTSTR. The maximum label length for the transformer-based PARSeq model is determined by the vocabulary used to create the synthetic dataset. We avoid using any data augmentation on synthetic datasets in accordance with community practise~\cite{baek2021if}.

Using a spatial transformer, the STARNet model transforms a resized input image of \(150 \times 18\) to \(100 \times 32\). Both STARNet and CRNN encoders accept images of size \(100 \times 32\), and the output feature maps are of size \(23 \times 1 \times 256\). All CRNN and STARNet models are trained on 2 Million synthetic images on a batch size of 32 and with ADADELTA~\cite{zeiler2012adadelta} optimizer for stochastic gradient descent. The number of epochs is fixed at 15. For each language, the models are tested on 0.5 million synthetic images.

We also run experiments on other Indic datasets, including MLT-17~\cite{nayef2017icdar2017} for Bengali (referred to as Bangla in MLT-17), MLT-19~\cite{nayef2019icdar2019} for Hindi, and Urdu-Text~\cite{chandio2020cursive} for Urdu. For all other public real datasets, we finetune on the train split and test our models on the test split (if no test splits are available, the proposer's val split is used).

\section{Result and Analysis}\label{result_analysis}

\subsection{Benchmarking}

In this section, we first list and compare the respective models' performance on a synthetic dataset, then on our proposed real dataset.

\subsubsection{Performance on Synthetic Dataset}

Table~\ref{table:synthNumbers} shows the CRR and WRR for each of the 13 languages achieved by various models. According to the data, the PARSeq model~\cite{bautista2022parseq} has clearly outperformed other models in terms of WRR and CRR in all 13 languages. This gain can be attributed to it being an attention-based model and using Permutation Language Modeling to further capture the context. Attention models have been shown to outperform CTC-based models in general by Latin solutions; the same holds true for the Indic language in the case of synthetic data points.

\begin{table}
\centering
\begin{threeparttable}
\caption{Performance on Synthetic Data}
\label{table:synthNumbers}
\begin{tabular}{p{2.5cm}|P{1.2cm}  P{1.2cm}|P{1.2cm}  P{1.2cm}|P{1.2cm}  P{1.2cm}}

\hline
\multirow{2}{*}{Language} & \multicolumn{2}{c|}{CRNN} &  \multicolumn{2}{c|}{STARNet}   & \multicolumn{2}{c}{PARSeq} \\ 

               & CRR & WRR & CRR & WRR & CRR & WRR    \\\hline\hline
Kannada         &  83.44    &  48.69   &  90.71  &  66.13   & \textbf{97.26}  & \textbf{87.92}  \\
Odiya           &    89.94  &  66.57   &  95.05  &  81.00   & \textbf{98.46}  & \textbf{93.27}  \\
Punjabi         &  88.88    &  66.11   &  93.86  &  78.34   & \textbf{97.08}  & \textbf{87.89}  \\
Urdu            &   76.83   &   39.90  &  86.51  &  58.35   & \textbf{95.26}  & \textbf{80.48}  \\
Marathi         &   86.91   &   59.87  &  93.55  &  76.32   & \textbf{99.08}  & \textbf{95.82}  \\
Assamese        &  88.86    &  67.90   &  94.76  &   82.93  & \textbf{99.21}  & \textbf{96.85}  \\
Manipuri (Meitei) &   89.83   &  67.57   &  94.71  &  80.74   & \textbf{98.79}  & \textbf{94.76}  \\
Malayalam       &    85.48  &  48.42   &  93.23  &  69.28   & \textbf{98.71}  & \textbf{92.27}  \\
Telugu          &   81.12   &  43.75   &  89.54  &   62.23  & \textbf{96.31}  & \textbf{83.4}   \\
Hindi\tnote{a}         &   89.83   &  73.15   &  95.78  &  83.93   & \textbf{99.13}  & \textbf{95.61}  \\
Bengali\tnote{a}         &   91.54   &  70.76   &  95.52  &  82.79   & \textbf{98.39}  & \textbf{92.56}  \\
Tamil\tnote{a}           &  82.86    &  48.19   &  95.40  &  79.90   & \textbf{97.88}  & \textbf{90.31}  \\
Gujarati\tnote{a}        &   94.43   &  81.85   &  97.80  &  91.40   & \textbf{98.82}  & \textbf{95.25}  \\
\hline
\end{tabular}
\begin{tablenotes}
\item [a] Values for CRNN and STARNet have been taken from Guna {\em et al.}~\cite{gunna2021transfer} as the training parameters and synthetic data generator were same.
\end{tablenotes}
\end{threeparttable}
\end{table}

\subsubsection{Performance on Real Dataset} 

\paragraph{\textbf{IndicSTR12 Dataset:}}

The CRR and WRR numbers for the three models on the IndicSTR12 dataset are listed in the Table~\ref{table:realNumbers}. Because the dataset is an extension of~\cite{gunna2021transfer} and IIIT-ILST~\cite{8270315}, the Hindi and Tamil numbers have been directly quoted from their work. We conducted separate experiments for Malayalam and Telugu (also covered by the two papers) because the cited works used a larger number of synthetic data to achieve higher accuracies. This was done to make a more accurate comparison with other languages' performance and to accurately gauge the models' performance on real data. Furthermore, the existing data for Gujarati was less than the required minimum of 1000 word-images per language, so this extension also supplements the Gujarati dataset.

A careful examination of the WRR and CRR numbers reveals that the PARSeq model outperforms in almost all cases where the real dataset is large enough, say greater than 1500. In a few cases, PARSeq falls short of the other two due to a lack of word-image instances for the model to train on.

\begin{table}[h]
\centering
\begin{threeparttable}
\caption{Performance on IndicSTR12}\label{table:realNumbers}
\begin{tabular}{p{2.5cm}|P{1.2cm}  P{1.2cm}|P{1.2cm}  P{1.2cm}|P{1.2cm}  P{1.2cm}|P {1.5cm}}
\hline
\multirow{2}{*}{Language} & \multicolumn{2}{c|}{CRNN} &  \multicolumn{2}{c|}{STARNet}   & \multicolumn{2}{c|}{PARSeq} & Word-images \\ 
                & CRR  & WRR & CRR     & WRR & CRR    & WRR &       \\
\hline 
\hline
Kannada         &    78.79  &   52.43  &  82.59   &  59.72  &    \textbf{88.64}  &  \textbf{63.57}   &   1074    \\
Odiya           &    80.39  &  54.74   &   86.97   &  66.30   &    \textbf{89.13}  &  \textbf{71.30}   &    3650   \\
Punjabi         &   83.15   &  68.85   &  84.93   &  62.5   &   \textbf{92.68}  &  \textbf{78.70}   &   3887    \\
Urdu            &    63.68  &  26.7   &   74.60   &  41.48   &   \textbf{76.97}  &  \textbf{44.19}   &    1375   \\
Marathi         &   70.79   & 50.96    &  83.73   &  58.65  &    \textbf{86.74}  &  \textbf{63.50}   &   1650    \\
Assamese        &   59.25   &  43.02   &  80.97   &  51.83  &    \textbf{81.36}  &   \textbf{52.70}  &   2154    \\
Malayalam       &    77.94  &  53.12   &  84.97  &  \textbf{70.09}   &    \textbf{90.10}  &  68.81   &   807    \\
Telugu          &    78.07  &  58.12   &  85.52  &  63.44   &    \textbf{92.18}  &  \textbf{71.94}   &    1211   \\
Hindi\tnote{a}    &   \textbf{78.84}   &  46.56   &  78.72  &  \textbf{46.60}   &    76.01    &   45.14  &   1150    \\
Bengali         &   59.86   &  48.21   &  80.26  &  57.70   &   \textbf{83.08}    &   \textbf{62.04}  &   3520    \\
Tamil\tnote{a}    &  75.05    &   59.06  &  \textbf{89.69}  &  \textbf{71.54}   &    87.56    &  67.35   &   2536    \\
Gujarati (New)\tnote{b}     &   52.22   &  23.05   &    \textbf{75.75}     &  41.80   &   74.49  &   \textbf{45.10}  & 1021  \\
Gujarati       &  53.34    &  42.58   & 74.82   &  51.56  &   \textbf{85.02}  &  \textbf{60.61}   &  922 \\
\hline
\end{tabular}
\begin{tablenotes}
\item [a] Values for CRNN and STARNet have been taken from Guna {\em et al.}~\cite{gunna2021transfer} as the training parameters and real data were same.
\item [b] Excluding~\cite{gunna2021transfer} data
\end{tablenotes}
\end{threeparttable}
\end{table}
\vspace{-0.01\textwidth}

\paragraph{\textbf{Other Public Dataset:}} 

\begin{figure}[h]
\centering
\includegraphics[width=\textwidth,keepaspectratio]{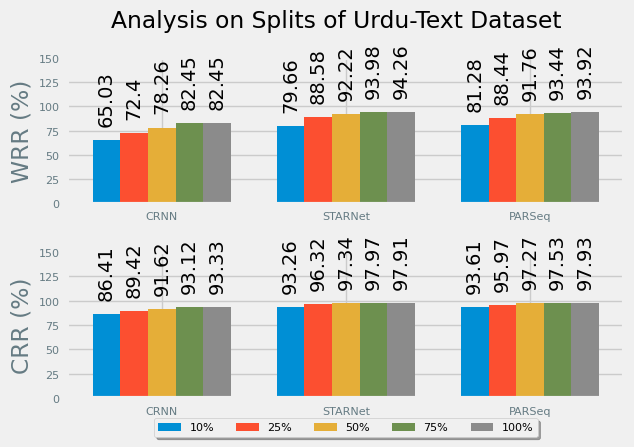}
\caption{Urdu Splits Analysis: Effect of training samples on STR models performance}
\label{fig:urduSplits}
\end{figure}

The models' overall performance, as shown in Table~\ref{table:otherDataset}, followed a similar pattern to that of the IndicSTR12 dataset. However, in contrast to trend seen for IndicSTR12, the performance of the STARNet models is comparable to that of PARSeq and somewhat better for MLT-19 Hindi and Urdu Text. Importantly, all models can be seen to achieve WRR substantially higher than in the case of the IndicSTR12 dataset if the number of word-images is similar. This trend is most pronounced for Urdu Text (refer Fig.~\ref{fig:urduSplits}, where 10\% of the original dataset roughly equals the number in IndicSTR12), MLT-19 Bengali, and MLT-17 Bengali.
This further demonstrates that IndicSTR12 is more challenging due to all the irregular samples and the fact that most of the images in the MLT-17, MLT-19, and Urdu Text dataset are frontal captures or regular in form. In the case of the Urdu Text dataset, the models achieved nearly identical performance utilizing just half of the dataset.

\begin{table}[h]
\centering
\begin{threeparttable}
\caption{Performance on Other Public Datasets}
\label{table:otherDataset}
\begin{tabular}{p{2.5cm}|P {2.0cm}|P{1.0cm}  P{1.0cm}|P{1.0cm}  P{1.0cm}|P{1.0cm}  P{1.0cm}} \hline
\multirow{2}{*}{Dataset} & Word-images & \multicolumn{2}{c|}{CRNN} &  \multicolumn{2}{c|}{STARNet}   & \multicolumn{2}{c}{PARSeq} \\ 
            & Train/Test  & CRR & WRR & CRR & WRR & CRR & WRR  \\\hline\hline
MLT-17 (Bengali) &   3237/713      &  79.98    &  55.30   &  85.24  &  65.73   & \textbf{88.72}  & \textbf{71.25}  \\
MLT-19 (Bengali)   &   3935/-    &    82.80  &  59.51   &  89.46  &  71.25   & \textbf{90.10}  & \textbf{72.59}  \\
MLT-19 (Hindi)  &   3931/-    &  86.48    &  67.90   &  91.00  &  \textbf{75.97}  & \textbf{91.80}  & 75.91 \\
Urdu -Text & 12076/1480           &   93.33  &   82.45  &  97.91  &  \textbf{94.26}   & \textbf{97.93}  & 93.92  \\
\hline
\end{tabular}
\end{threeparttable}
\end{table}

\subsubsection{Error Analysis}

\begin{figure}[h]
     \centering
     \begin{subfigure}[c]{0.4\textwidth}
         \centering
         \includegraphics[width=\textwidth,keepaspectratio]{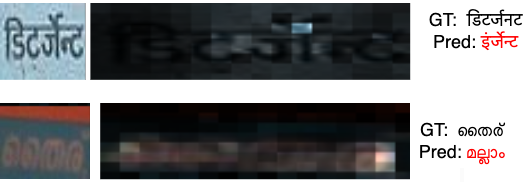}
         \caption{Low-resolution Images}
         \label{fig:pred_low_res}
     \end{subfigure}
     \hfill
     \begin{subfigure}[c]{0.4\textwidth}
         \centering
         \includegraphics[width=\textwidth,keepaspectratio]{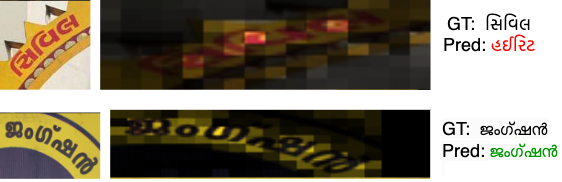}
         \caption{Rotated and Curved Text}
         \label{fig:pred_curved}
     \end{subfigure}

     \begin{subfigure}[c]{0.4\textwidth}
         \centering
         \includegraphics[width=\textwidth,keepaspectratio]{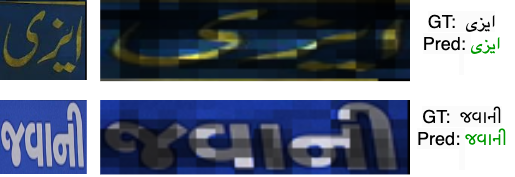}
         \caption{Iconic Text: Urdu and Gujarati}
         \label{fig:pred_icon_urdu_guj}
     \end{subfigure}
     \hfill
     \begin{subfigure}[c]{0.4\textwidth}
         \centering
         \includegraphics[width=\textwidth]{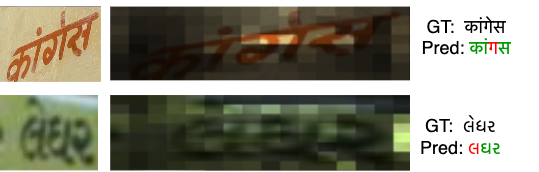}
         \caption{Inadequate Attention to Matras}
         \label{fig:pred_matra_miss}
     \end{subfigure}

     \begin{subfigure}[c]{0.4\textwidth}
         \centering
         \includegraphics[width=\textwidth]{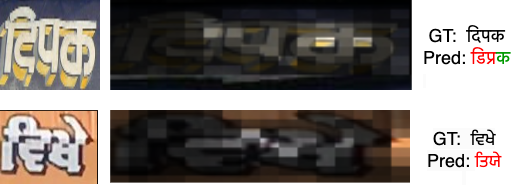}
         \caption{Difficult Fonts and Distorted text examples}
         \label{fig:incorrect_font_shadow}
     \end{subfigure}
     \hfill
     \begin{subfigure}[c]{0.4\textwidth}
         \centering
         \includegraphics[width=\textwidth]{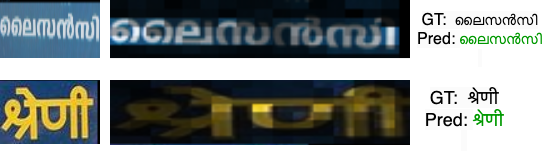}
         \caption{Iconic Text: Hindi and Malayalam}
         \label{fig:pred_icon_hindi_mal}
     \end{subfigure}
        \caption{Error Analysis using Attention Maps}
        \label{fig:vit_atten_analysis}
\end{figure}

Some failure cases were found after analyzing the PARSeq model's ViT encoder attention maps for prediction mistakes (Fig. ~(\ref{fig:vit_atten_analysis})) using the method proposed by~\cite{abnar2020quantifying}. The PARSeq model can recognize the textual region even in irregular word-image examples , but its predictions for the text are severely inaccurate for low-resolution images (Fig.~(\ref{fig:pred_low_res})) and only somewhat reliable for rotated or curved text (Fig. ~(\ref{fig:pred_curved})). Additionally, because lower and upper matra are not given adequate consideration, the model misses a lot of accurate predictions (Fig.~(\ref{fig:pred_matra_miss})). Another notable instance of failure is when distorted text or unusual fonts cause shadows or border thickness to be seen as a different character (Fig.~(\ref{fig:incorrect_font_shadow})). The model does reasonably well when dealing with typical iconic texts (Fig.~(\ref{fig:pred_icon_hindi_mal})) and does well overall when dealing with lengthy texts.

\subsection{Multi-lingual Training}

\begin{table}[!ht]
\centering
\begin{threeparttable}
\caption{Multi-Lingual Training}
\label{table:multiLingual}
\begin{tabular}{p{3.0 cm}|P {3.0cm}|P{1.2cm}  P{1.2cm}}\hline
\multirow{2}{*}{Lang.(s) Trained On} & \multirow{2}{*}{Lang. Tested On} & \multicolumn{2}{c}{PARSeq} \\ 
            & & CRR & WRR \\\hline \hline
Hindi &  Hindi    &  76.01   & 45.14  \\
Hindi-Gujarati    &   Hindi   &    \textbf{76.41}  & \textbf{49.65} \\
\hline
Gujarati  &   Gujarati    &  74.49    &  45.51 \\
Hindi-Gujarati    &   Gujarati  &   \textbf{75.68}  &   \textbf{45.70} \\
\hline
\end{tabular}
\end{threeparttable}
\end{table}

The community has investigated transfer learning for STR models~\cite{gunna2021transfer} as well as multi-lingual models in the case of OCR~\cite{mathew2016multilingual} since there few real training examples available. It is warranted as Indic language groups share a syntactic and semantic commonality. We looked into the multi-lingual model perspective here for Indic Languages in STR to provide a demonstrative example. In order to have a fair comparison, since single language models are trained on 2M synthetic datasets and finetuned on their respective real images, we trained our multi-lingual model on the same number of synthetic images—1M Hindi and 1M Gujarati—and then finetuned on a combined Hindi-Gujarati real images dataset, to demonstrates that multi-lingual approach does indeed aid model in learning each individual language better. Results~\ref{table:multiLingual} indicate a 4.0\% increase in WRR for Hindi and a 0.20\% increase for Gujarati.

\section{Conclusion}

We assembled a real dataset for STR in Indian languages. IndicSTR12 is the most comprehensive dataset available, and it is a first in many languages. We generated 3 million synthetic word-images for all 13 languages in addition to real datasets for faster STR solution development for Indic languages. In addition to benchmarking STR performance on three models, this paper establishes the need for even more data to leverage the learning powers of SOTA Latin models. Because Indian scripts are more complex than Latin scripts, they necessitate a more comprehensive training resource. In the future, this large dataset in both the real and synthetic domains will aid the Indic Scene text community in developing solutions for Indic STR that are on par with Latin solutions.
\label{conclusion_paper}

\section*{Acknowledgement}
This work is supported by MeitY, Government of India, through the NLTM-Bhashini project.

\bibliographystyle{splncs04}
\bibliography{egbib}

\end{document}